# Risk Bounds for Infinitely Divisible Distribution

**Chao Zhang**
CeMNet Annexe
School of Computer Engineering
Nanyang Technological University
639798, Singapore.

**Dacheng Tao**
Centre for Quantum Computation
& Intelligent Systems (QCIS), FEIT
University of Technology, Sydney
Broadway, NSW 2007, Australia.

## Abstract

In this paper, we study the risk bounds for samples independently drawn from an infinitely divisible (ID) distribution. In particular, based on a martingale method, we develop two deviation inequalities for a sequence of random variables of an ID distribution with *zero* Gaussian component. By applying the deviation inequalities, we obtain the risk bounds based on the covering number for the ID distribution. Finally, we analyze the asymptotic convergence of the risk bound derived from one of the two deviation inequalities and show that the convergence rate of the bound is faster than the result for the generic i.i.d. empirical process (Mendelson, 2003).

## 1 Introduction

A probability distribution is infinitely divisible if and only if it can be represented as the distribution of the sum of an arbitrary number of independently and identically distributed (i.i.d.) random variables. Infinitely divisible (ID) distribution covers a large body of probability distributions, *e.g.*, Poisson, geometric, log-normal, noncentral chi-square, exponential, Gamma, Pareto and Cauchy (Bose *et al.*, 2001). It is essential to study ID distribution and its statistical properties, because many practical problems can be reduced to an empirical learning process related to ID distributions. For example, ID distributions have been used to handle the option valuation (Heston, 2004) and the evaluation to the credit risk (Moosbrucker, 2007)) in finance. Moreover, it has been well acknowledged that the distributions of natural image statistics also have the infinite divisibility (Mumford and Gidas, 2001; Chainais, 2007). It is necessary to analyze the asymptotical convergence of the learning process, when the number of samples goes to the *infinity.* The risk bound is the main method to study the asymptotical behavior.

Let $\mathcal{Z} := (\mathcal{X}, \mathcal{Y}) \subseteq \mathbb{R}^K$ be a space with $K = I + J$, where $\mathcal{X} \subseteq \mathbb{R}^I$ is the input space and $\mathcal{Y} \subseteq \mathbb{R}^J$ is the corresponding output space. We would like to find a function $g^* : \mathcal{X} \to \mathcal{Y}$ such that for any $\mathbf{x} \in \mathcal{X}$, $g^*(\mathbf{x})$ can precisely estimate the output $\mathbf{y} \in \mathcal{Y}$. This can be achieved by minimizing the expected risk

$$\mathrm{E}(\ell(g(\mathbf{x}), \mathbf{y})) := \int \ell(g(\mathbf{x}), \mathbf{y}) dP(\mathbf{z}), \tag{1}$$

where $\ell : \mathcal{Y}^2 \to \mathbb{R}$ is a loss function and $P(\mathbf{z})$ stands for the distribution of $\mathbf{z} = (\mathbf{x}, \mathbf{y}) \in \mathcal{Z}$. Since the distribution $P(\mathbf{z})$ is unknown, the target $g^*$ usually cannot be directly obtained by minimizing (1). Instead, the empirical risk minimization principle can be used to handle this issue. We consider a function class $\mathcal{G}$ composed of real-valued functions defined on $\mathcal{Z}$ and a sample set $\mathbf{Z}_1^N := \{\mathbf{z}_n\}_{n=1}^N \subset \mathcal{Z}$ with $\mathbf{z}_n = (\mathbf{x}_n, \mathbf{y}_n)$. Given a function $g \in \mathcal{G}$, the empirical risk is defined by

$$\mathrm{E}_N(\ell(g(\mathbf{x}), \mathbf{y})) := \frac{1}{N} \sum_{n=1}^{N} \ell(g(\mathbf{x}_n), \mathbf{y}_n), \tag{2}$$

which is regarded as an approximation of the expected risk (1). Alternatively, we minimize the empirical risk to obtain an estimate to $g^*$. We then define the loss function class

$$\mathcal{F} := \{\mathbf{z} \mapsto \ell(g(\mathbf{x}), \mathbf{y}) : g \in \mathcal{G}\}.$$

To simplify the notation, for any $f \in \mathcal{F}$, we define

$$\mathrm{E}f := \int f(\mathbf{z}) dP(\mathbf{z}) \quad \text{and} \quad \mathrm{E}_N f := \frac{1}{N} \sum_{n=1}^{N} f(\mathbf{z}_n).$$

The upper bound of $\sup_{f \in \mathcal{F}} \left| \mathrm{E}_N f - \mathrm{E}f \right|$ is called the risk bound and plays an important role in statistical learning theory. The risk bound measures the probability that a function obtained by an algorithm has a sufficiently small error. In general, in order to obtain the risk bound, one has to consider the following three aspects: complexity measures of function classes, deviation (or concentration) inequalities and symmetrization inequalities.

There have been some risk bounds proposed for the generic i.i.d. empirical process. For example, Vaart and Wellner (1996) showed the risk bounds based on the Rademacher complexity and the covering number. Vapnik (1999) gave risk bounds based on the annealed VC entropy and the VC dimension, respectively. Bartlett *et al.* (2005) developed the local Rademacher complexity and obtained a sharp error bound for a particular function class $\{f \in \mathcal{F} | \mathrm{E}f^2 < \beta \mathrm{E}f, \beta > 0\}$. Moreover, Mohri and Rostamizadeh (2008) studied the risk bound based on the Rademacher complexity for stationary $\beta$-mixing sequences.

In this paper, we study the risk bounds for samples independently drawn from an ID distribution with *zero* Gaussian component. Although it is a special case of the generic i.i.d. empirical process, it is still important to study risk bounds for the following reasons:

- ID distribution covers a large body of probability distributions and many practical problems can be reduced to an empirical learning process related to ID distribution.
- In order to obtain the desired risk bounds, we have to develop new deviation inequalities and they are different from those for the generic i.i.d. empirical process.
- We will show that the convergence rate of the risk bound reaches $O\big(\big(\frac{\ln \mathrm{E}\{\mathcal{N}(\mathcal{F},\xi,L_1(\mathbf{Z}_1^{2N}))\}}{N}\big)^{\frac{1}{1.3}}\big)$, where $\mathcal{N}$ stands for the covering number. It is faster than $O\big(\big(\frac{\ln \mathrm{E}\{\mathcal{N}(\mathcal{F},\xi,L_1(\mathbf{Z}_1^{N}))\}}{N}\big)^{\frac{1}{2}}\big)$ for the generic i.i.d. empirical process (*cf.* Theorem 2.3 in (Mendelson, 2003)).

In order to obtain the desired risk bounds, it is necessary to obtain suitable concentration (or deviation) inequalities. The deviation inequalities have been widely used to study the consistence of the empirical risk minimization principle and they measure the difference between the expected risk and the empirical risk. Houdré (2002) has proposed deviation inequalities for ID distribution with *zero* Gaussian component. However, his results are only valid for one random variable and thus his results cannot be directly applied to a sequence of ID random variables.

Based on a martingale method, we generalize Houdré's results (Houdré, 2002) and obtain two deviation inequalities for a sequence of ID random variables with *zero* Gaussian component. By using the deviation inequalities, we then obtain the risk bounds based on the covering number for ID distribution. Finally, we analyze the asymptotic convergence of the risk bound derived from the deviation inequality (10) and show that the bound can provide a faster the convergence rate than the result for the generic i.i.d. empirical process (Mendelson, 2003).

The rest of this paper is organized as follows. Section 2 introduces ID distribution. Two deviation inequalities are presented in Section 3. We give the risk bounds for ID distribution in Section 4. In Section 5, we analyze the asymptotic convergence of the risk bound. Some proofs of main results are shown in Section 6 and the last section concludes the paper.

## 2 Infinitely Divisible Distributions

This section introduces some preliminaries about infinitely divisible (ID) distribution and details are given in (Sato, 2004). The ID distribution can be defined by using the characteristic function as follows:

**Definition 2.1** *Let $\phi(t)$ be the characteristic function of a random variable $\mathbf{z}$*

$$\phi(t) := \mathrm{E}\left\{\mathrm{e}^{it\mathbf{z}}\right\} = \int_{-\infty}^{+\infty} \mathrm{e}^{it\mathbf{z}} dP(\mathbf{z}). \tag{3}$$

*The distribution of $\mathbf{z}$ is infinitely divisible if and only if for any $N \in \mathbb{N}$, there exists a characteristic function $\phi_N(t)$ such that*

$$\phi(t) = \underbrace{\phi_N(t) \times \cdots \times \phi_N(t)}_{N}, \tag{4}$$

*where "$\times$" stands for multiplication.*

According to the definition, a random variable has the infinite divisibility if and only if it can be represented as the sum of an arbitrary number of i.i.d. random variables. We then introduce the Lévy measure and show the characteristic exponent of an ID distribution (Sato, 2004).

**Definition 2.2** *Let $\nu$ be a Borel measure defined on $\mathbb{R}^K/\{0\}$. Then, the $\nu$ is said to be a Lévy measure if*

$$\int_{\mathbb{R}^K/\{0\}} \min\{\|\mathbf{u}\|^2, 1\} \nu(d\mathbf{u}) < \infty, \tag{5}$$

*and $\nu(0) = 0$.*

The Lévy measure describes the expected number of a certain height jump in a time interval of unit length. The characteristic exponent of an ID random variable is given by the following theorem (Sato, 2004).

**Theorem 2.3** *(Lévy-Khintchine) A Borel probability measure $\mu$ of a random variable $\mathbf{z} \in \mathbb{R}^K$ is infinitely divisible if and only if there exists a triplet $(\mathbf{a}, \Sigma, \nu)$ such that for all $\theta \in \mathbb{R}^K$, the characteristic exponent $\ln \phi_\mu$ is of the form*

$$\ln\phi_\mu(\theta) = i\langle \mathbf{a}, \theta\rangle - \frac{1}{2}\langle \theta, \Sigma\theta\rangle + \int_{\mathbb{R}^K/\{0\}} \left(\mathrm{e}^{i\langle\theta,u\rangle} - 1 - i\langle\theta, \mathbf{u}\rangle \mathbf{1}_{\|\mathbf{u}\|\leq 1}\right)\nu(d\mathbf{u}), \quad (6)$$

*where $\mathbf{a} \in \mathbb{R}^K$, $\Sigma$ is a $K \times K$ positive-definite symmetric matrix, $\nu$ is a Lévy measure on $\mathbb{R}^K/\{0\}$, and $\langle \cdot, \cdot\rangle$ and $\|\cdot\|$ stand for the inner product and a norm in $\mathbb{R}^K$, respectively.*

Theorem 2.3 shows that an ID distribution can be completely determined by a triplet $(\mathbf{a}, \Sigma, \nu)$, where "$\mathbf{a}$" is the drift of a Brownian motion, "$\Sigma$" is the Gaussian component and "$\nu$" is a Lévy measure. Thus, $(\mathbf{a}, \Sigma, \nu)$ is termed as the generating triplet of an ID distribution.

## 3 Deviation Inequality

Houdré (2002) gave deviation inequalities for one single ID random variable with the generating triplet $(\mathbf{a}, \mathbf{0}, \nu)$. However, his results are not applicable to the case of a sequence of ID random variables. In this section, we utilize a martingale method to generalize Houdré's results and obtain two derivation inequalities for a sequence of ID random variables with $(\mathbf{a}, \mathbf{0}, \nu)$.

**Theorem 3.1** *Assume that $f$ is a $\lambda$-Lipschitz continuous function. Let $\mathbf{Z}_1^N = \{\mathbf{z}_n\}_{n=1}^N$ be a sample set independently drawn from an ID distribution $\mathcal{Z}$ with $(\mathbf{a}, \mathbf{0}, \nu)$. If $\mathrm{E}\mathrm{e}^{t\|\mathbf{z}\|} < +\infty$ holds for some $t > 0$, then we have for all $0 < \xi < \tau\big((M/\lambda)^-\big)$,*

$$\Pr\left\{\left|F\left(\mathbf{Z}_1^N\right) - \mathrm{E}F\right| > \xi\right\} \leq \exp\left\{-\int_0^\xi \tau^{-1}(s)ds\right\}, \quad (7)$$

*where*

$$F\left(\mathbf{Z}_1^N\right) := \sum_{n=1}^N f(\mathbf{z}_n), \quad (8)$$

*$\tau(a^-)$ is the left-hand limit of $\tau$ at $a$, $M = \sup\{t \geq 0 : \mathrm{E}\mathrm{e}^{t\|\mathbf{z}\|} < +\infty\}$ and $\tau^{-1}$ is the inverse of*

$$\tau(t) = N\int_{\mathbb{R}^K} \lambda\|\mathbf{u}\|\left(\mathrm{e}^{t\lambda\|\mathbf{u}\|} - 1\right)\nu(d\mathbf{u}), \quad (9)$$

*with the domain of $0 < t < M/\lambda$.*

Because of an integral of $\tau^{-1}$, the deviation inequality (7) cannot be directly used to study the asymptotic behavior of $\Pr\left\{\left|F\left(\mathbf{Z}_1^N\right) - \mathrm{E}F\right| > \xi\right\}$, when $N$ goes to the *infinity*. Thus, we develop the other deviation inequality by introducing an extra condition that the Lévy measure $\nu$ has a bounded support.

**Corollary 3.2** *Following notations in Theorem 3.1, let $V = \int_{\mathbb{R}^K}\|\mathbf{u}\|^2\nu(d\mathbf{u})$. If $\nu$ has a bounded support with $R = \inf\{\rho > 0 : \nu(\{\mathbf{u} : \|\mathbf{u}\| > \rho\}) = 0\}$, then we have for any $\xi > 0$,*

$$\Pr\left\{\left|F\left(\mathbf{Z}_1^N\right) - \mathrm{E}F\right| > \xi\right\} \leq \exp\left\{\frac{\xi}{\lambda R} - \left(\frac{\xi}{\lambda R} + \frac{NV}{(\lambda R)^2}\right)\ln\left(1 + \frac{\xi\lambda R}{NV}\right)\right\}. \quad (10)$$

Based on the above two deviation inequalities, we can obtain risk bounds for an ID distribution.

## 4 Risk Bound for Infinitely Divisible Distribution

In this section, we present the risk bounds based on the covering number for ID distribution with the generating triple $(\mathbf{a}, \mathbf{0}, \nu)$. The covering number of $\mathcal{F}$ is defined as follows and we refer to (Mendelson, 2003) for details.

**Definition 4.1** *Let $\mathbf{Z}_1^N = \{\mathbf{z}_n\}_{n=1}^N$ be a sample set. For any $1 \leq p \leq \infty$ and $\xi > 0$, the covering number of $\mathcal{F}$ at radius $\xi$, with respect to $L_p(\mathbf{Z}_1^N)$, denoted by $\mathcal{N}(\mathcal{F}, \xi, L_p(\mathbf{Z}_1^N))$ is the minimum size of a cover of radius $\xi$.*

Subsequently, we come up with the following theorems, which are the main results of this paper.

**Theorem 4.2** *Assume that $\mathcal{F}$ is a function class composed of $\lambda$-Lipschitz continuous functions with the range $[A, B]$. Let $\mathbf{Z}_1^N = \{\mathbf{z}_n\}_{n=1}^N$ be a sample set independently drawn from an ID distribution with $(\mathbf{a}, \mathbf{0}, \nu)$. If $\mathrm{E}\mathrm{e}^{t\|\mathbf{z}\|} < +\infty$ holds for some $t > 0$, then for all $\xi > 0$ such that $0 < N\xi/8 < \tau\big((M/\lambda)^-\big)$ and $N\xi^2 \geq 32\max\{A^2, B^2\}$, we have*

$$\Pr\left\{\sup_{f\in\mathcal{F}}\left|\mathrm{E}_N f - \mathrm{E}f\right| > \xi\right\} \leq 2\mathrm{E}\left\{\mathcal{N}\left(\mathcal{F}, \frac{\xi}{8}, L_1(\mathbf{Z}_1^{2N})\right)\right\}\exp\left\{-\int_0^{\frac{N\xi}{8}} \tau^{-1}(s)ds\right\}, \quad (11)$$

*where $\tau(a^-)$ denotes the left-hand limit of $\tau$ at the point $a$, $M = \sup\left\{t \geq 0 : \mathrm{E}\mathrm{e}^{t\|\mathbf{z}\|} < +\infty\right\}$ and $\tau^{-1}$ is the inverse of*

$$\tau(t) = N\int_{\mathbb{R}^K} \lambda\|\mathbf{u}\|\left(\mathrm{e}^{t\lambda\|\mathbf{u}\|} - 1\right)\nu(d\mathbf{u}),$$

*with the domain of $0 < t < M/\lambda$.*

Since (11) is given by incorporating the integrals of $\tau^{-1}$, the asymptotic behavior of the risk bounds cannot be explicitly reflected, when $N$ goes to the *infinity*. Moreover, the applicability of Theorem 4.2 is restricted by the two conditions $0 < N\xi/8 < \tau\big((M/\lambda)^-\big)$ and $N\xi^2 \geq 32\max\{A^2, B^2\}$. To overcome this limitation, we develop another risk bound for ID distribution by adding a mild condition that the Lévy measure $\nu$ has a bounded support, where we retain the condition $N\xi^2 \geq 32\max\{A^2, B^2\}$ but remove the condition $0 < N\xi/8 < \tau\big((M/\lambda)^-\big)$. Its proof is similar to that of Theorem 4.2, so we omit it.

**Theorem 4.3** *Following notations in Theorem 4.2, let* $V = \int_{\mathbb{R}^K} \|\mathbf{u}\|^2\nu(d\mathbf{u})$. *If* $\nu$ *has a bounded support with* $R = \inf\{\rho > 0 : \nu(\{\mathbf{u} : \|\mathbf{u}\| > \rho\}) = 0\}$*, then we have for any* $\xi > 0$ *such that* $N\xi^2 \geq 32\max\{A^2, B^2\}$*,*

$$\Pr\left\{\sup_{f\in\mathcal{F}} \big|\mathrm{E}_N f - \mathrm{E}f\big| > \xi\right\} \tag{12}$$
$$\leq 2\mathrm{E}\left\{\mathcal{N}\left(\mathcal{F}, \frac{\xi}{8}, L_1(\mathbf{Z}_1^{2N})\right)\right\}\exp\left\{\frac{NV}{\lambda^2R^2}\Gamma\left(\frac{\xi\lambda R}{8V}\right)\right\},$$

*where*

$$\Gamma(x) = x - (x+1)\ln(x+1). \tag{13}$$

Theorem 4.3 shows that if the Lévy measure $\nu$ has a bounded support, we can obtain a risk bound that can explicitly reflect its asymptotic behavior, when $N$ goes to the *infinity*. In the next section, we analyze the asymptotic convergence of the risk bound.

## 5 Asymptotic Convergence

In this section, we study the asymptotic convergence of the risk bound (12) and show that if samples are independently drawn from an ID distribution, $\sup_{f\in\mathcal{F}} \big|\mathrm{E}_N f - \mathrm{E}f\big|$ has a faster convergence rate than the result for the generic i.i.d. empirical process. First, we give a convergence result which can be directly obtained from (12) and (13).

**Theorem 5.1** *Following notations in Theorem 4.3, let* $x^*$ *be the solution of the equation*

$$\Gamma(x) = x - (x+1)\ln(x+1) = 0, \;\; x > 0.$$

*If there holds that*

$$\lim_{N\to\infty} \frac{\ln \mathrm{E}\left\{\mathcal{N}\left(\mathcal{F}, \xi/8, L_1(\mathbf{Z}_1^{2N})\right)\right\}}{N} = 0, \tag{14}$$

*then we have for any* $\xi > \frac{8x^*V}{\lambda R}$ *such that* $N\xi^2 \geq 32\max\{A^2, B^2\}$*,*

$$\lim_{N\to\infty} \Pr\left\{\sup_{f\in\mathcal{F}} \big|\mathrm{E}_N f - \mathrm{E}f\big| > \xi\right\} = 0.$$

As shown in Theorem 5.1, if the condition (14) is valid, there holds that for some $\xi > 0$, the left hand side of (12) converges to *zero*, when the number of samples goes to the *infinity*. This is in accordance with the convergence result for the generic i.i.d. empirical process derived from Theorem 2.3 in (Mendelson, 2003). However, because of the particularity of ID distribution, the convergence rate of $\sup_{f\in\mathcal{F}} |\mathrm{E}_N f - \mathrm{E}f|$ will be different from the case of the generic i.i.d. empirical process. Next, we give the detailed discussion on the convergence rate.

Given a number $\widetilde{x} > 1$, consider the following equation with respect to $\gamma > 0$

$$\widetilde{x} - (\widetilde{x}+1)\ln(\widetilde{x}+1) = -\widetilde{x}^{\gamma}, \tag{15}$$

and denote its solution as

$$\gamma(\widetilde{x}) := \frac{\ln\big((\widetilde{x}+1)\ln(\widetilde{x}+1) - \widetilde{x}\big)}{\ln(\widetilde{x})}. \tag{16}$$

Then, we have for any $0 < \widetilde{\gamma} \leq \gamma(\widetilde{x})$,

$$\widetilde{x} - (\widetilde{x}+1)\ln(\widetilde{x}+1) \leq -\widetilde{x}^{\widetilde{\gamma}}. \tag{17}$$

By using (17), we can represent the upper bound of $\sup_{f\in\mathcal{F}} \big|\mathrm{E}_N f - \mathrm{E}f\big|$ as follows.

**Theorem 5.2** *Follow notations in Theorem 4.3. For any* $\xi > 0$ *such that* $N\xi^2 \geq 32\max\{A^2, B^2\}$ *and* $\frac{\xi\lambda R}{8V} > 1$*, letting*

$$\epsilon = 2\mathrm{E}\left\{\mathcal{N}\left(\mathcal{F}, \frac{\xi}{8}, L_1(\mathbf{Z}_1^{2N})\right)\right\}\exp\left\{\frac{NV}{\lambda^2R^2}\Gamma\left(\frac{\xi\lambda R}{8V}\right)\right\},$$

*we have with probability at least* $1-\epsilon$*,*

$$\sup_{f\in\mathcal{F}} \big|\mathrm{E}_N f - \mathrm{E}f\big|$$
$$\leq \left(\frac{8\lambda R\left(\ln \mathrm{E}\left\{\mathcal{N}\left(\mathcal{F}, \xi/8, L_1(\mathbf{Z}_1^{2N})\right)\right\} - \ln(\epsilon/2)\right)}{N\left(\frac{\lambda R}{8V}\right)^{\gamma-1}}\right)^{\frac{1}{\gamma}},$$

*where* $0 < \gamma \leq \gamma\left(\frac{\xi\lambda R}{8V}\right)$.

**Proof.** It can be directly resulted from the combination of (12), (15), (16) and (17). ■

The above theorem shows that with probability at least $1-\epsilon$,

$$\sup_{f\in\mathcal{F}} \big|\mathrm{E}_N f - \mathrm{E}f\big| \tag{18}$$
$$\leq O\left(\left(\frac{\ln \mathrm{E}\left\{\mathcal{N}\left(\mathcal{F}, \xi/8, L_1(\mathbf{Z}_1^{2N})\right)\right\} - \ln(\epsilon/2)}{N}\right)^{\frac{1}{\gamma}}\right).$$

In order to find the convergence rate of $\sup_{f\in\mathcal{F}} |\mathrm{E}_N f - \mathrm{E}f|$, we have to study the upper bound of the function

$\gamma(x)$ $(x > 1)$. According to (16), for any $x > 1$, we consider the derivative of $\gamma(x)$

$$\gamma'(x) = \frac{\ln(x+1)}{\ln(x)\big((x+1)\ln(x+1) - x\big)} - \frac{\ln\big((x+1)\ln(x+1) - x\big)}{x(\ln x)^2}, \tag{19}$$

and draw the function curve of $\gamma'(x)$ in Fig. 1.

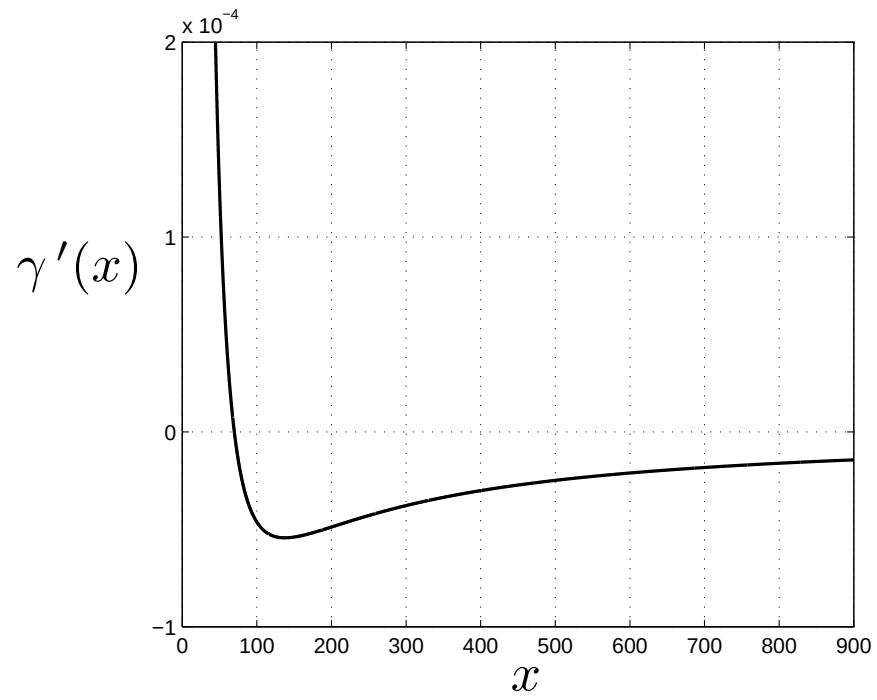

Figure 1: The Function Curve of $\gamma'(x)$

As shown in Fig. 1, there is only one solution to the equation $\gamma'(x) = 0$ $(x > 1)$. Letting the solution be $\widehat{x}$, we then have $\gamma'(x) > 0$ $(1 < x < \widehat{x})$ and $\gamma'(x) < 0$ $(x > \widehat{x})$. Meanwhile, by (19), there holds that

$$\lim_{x \to +\infty} \gamma'(x) = 0. \tag{20}$$

Therefore, we obtain that

$$\widehat{x} = \arg\max_{x>1} \gamma(x). \tag{21}$$

Our further numerical experiment shows that the value of $\widehat{x}$ approximately equals to 69.85 and the maximum of $\gamma(x)$ $(x > 1)$ is not larger than 1.3 (*cf.* Fig. 2). Thus, according to (18), we can obtain that with probability at least $1 - \epsilon$,

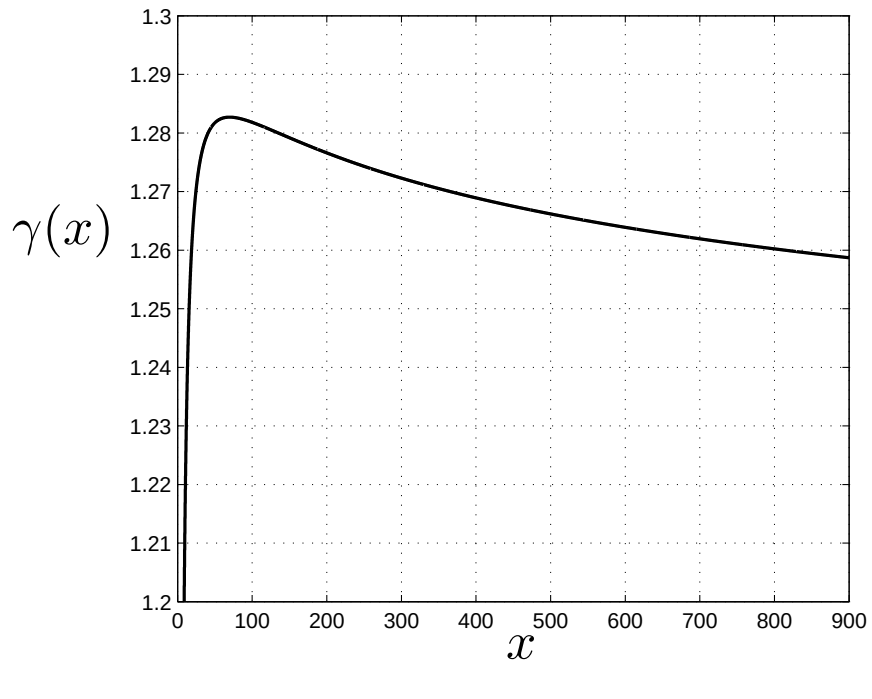

Figure 2: The Function Curve of $\gamma(x)$

$$\sup_{f\in\mathcal{F}} \big|\mathrm{E}_N f - \mathrm{E} f\big| \tag{22}$$
$$\leq O\left(\left(\frac{\ln \mathrm{E}\left\{\mathcal{N}\left(\mathcal{F}, \xi/8, L_1(\mathbf{Z}_1^{2N})\right)\right\} - \ln(\epsilon/2)}{N}\right)^{\frac{1}{1.3}}\right).$$

Compared with the result of the generic i.i.d. empirical process given in Theorem 2.3 of (Mendelson, 2003)

$$\sup_{f\in\mathcal{F}} \big|\mathrm{E}_N f - \mathrm{E} f\big|$$
$$\leq O\left(\left(\frac{\ln \mathrm{E}\left\{\mathcal{N}\left(\mathcal{F}, \xi, L_1(\mathbf{Z}_1^{N})\right)\right\}}{N}\right)^{\frac{1}{2}}\right),$$

we find that if samples are independently drawn from an ID distribution with *zero* Gaussian component, $\sup_{f\in\mathcal{F}} \big|\mathrm{E}_N f - \mathrm{E} f\big|$ has a faster convergence rate.

# 6 Proofs of Main Results

In this section, we prove Theorem 3.1, Corollary 3.2 and Theorem 3.1, respectively.

## 6.1 Martingale Method

In this paper, we use the following martingale method to generalize Houdré's deviation inequalities (Houdré, 2002) to a sequence of ID random variables.

For any $0 \leq m \leq N$, define a random variable

$$S_m := \mathrm{E}\left\{F(\mathbf{Z}_{n=1}^N)|\mathbf{Z}_1^m\right\}, \tag{23}$$

where $\mathbf{Z}_1^m = \{\mathbf{z}_1, \cdots, \mathbf{z}_m\} \subseteq \mathbf{Z}_1^N$ and $\mathbf{Z}_1^0 = \varnothing$. It is direct that $S_0 = \mathrm{E}F$ and $S_N = F(\mathbf{Z}_1^N)$.

According to (8) and (23), for any $1 \leq m \leq N$, letting

$$\psi_m(\mathbf{Z}_1^N) := S_m - S_{m-1}, \tag{24}$$

we have

$$\begin{aligned}\psi_m(\mathbf{Z}_1^N) =& \mathrm{E}\left\{F(\mathbf{Z}_1^N)|\mathbf{Z}_1^m\right\} - \mathrm{E}\left\{F(\mathbf{Z}_1^N)|\mathbf{Z}_1^{m-1}\right\} \\ =& \mathrm{E}\left\{\sum_{n=1}^N f(\mathbf{z}_n)\Big|\mathbf{Z}_1^m\right\} - \mathrm{E}\left\{\sum_{n=1}^N f(\mathbf{z}_n)\Big|\mathbf{Z}_1^{m-1}\right\} \\ =& \sum_{n=1}^m f(\mathbf{z}_n) + \mathrm{E}\left\{\sum_{n=m+1}^N f(\mathbf{z}_n)\right\} \\ &- \left(\sum_{n=1}^{m-1} f(\mathbf{z}_n) + \mathrm{E}\left\{\sum_{n=m}^N f(\mathbf{z}_n)\right\}\right) \\ =& f(\mathbf{z}_m) - \mathrm{E}f(\mathbf{z}_m),\end{aligned} \tag{25}$$

and thus

$$\mathrm{E}\left\{\psi_m(\mathbf{Z}_1^N)\big|\mathbf{Z}_1^{m-1}\right\} = \mathrm{E}\left\{\psi_m(\mathbf{Z}_1^N)\right\} = 0. \tag{26}$$

Moreover, we also have the following lemma.

**Lemma 6.1** *Following the notation in* (8) *and* (24), *we have*

$$\Pr\left\{F\left(\mathbf{Z}_1^N\right)-\mathrm{E}F>\xi\right\}\leq \mathrm{e}^{-t\xi}\prod_{m=1}^{N}\mathrm{E}\left\{\mathrm{e}^{t\psi_m}\right\}. \quad (27)$$

**Proof.** According to (25), Markov's inequality and the law of iterated expectation, we have

$$\begin{aligned}
&\Pr\left\{F\left(\mathbf{Z}_1^N\right)-\mathrm{E}F>\xi\right\}\\
\leq&\mathrm{e}^{-t\xi}\mathrm{E}\left\{\mathrm{e}^{t\left(F\left(\mathbf{Z}_1^N\right)-\mathrm{E}F\right)}\right\}\\
=&\mathrm{e}^{-t\xi}\mathrm{E}\left\{\mathrm{e}^{t\sum_{n=1}^{N}(S_m-S_{m-1})}\right\}\\
=&\mathrm{e}^{-t\xi}\mathrm{E}\left\{\mathrm{E}\left\{\mathrm{e}^{t\sum_{m=1}^{N}(S_m-S_{m-1})}\Big|\mathbf{Z}_1^{N-1}\right\}\right\}\\
=&\mathrm{e}^{-t\xi}\mathrm{E}\left\{\mathrm{e}^{t\sum_{m=1}^{N-1}(S_m-S_{m-1})}\mathrm{E}\left\{\mathrm{e}^{t(S_N-S_{N-1})}\Big|\mathbf{Z}_1^{N-1}\right\}\right\}\\
=&\mathrm{e}^{-t\xi}\prod_{n=1}^{N}\mathrm{E}\left\{\mathrm{e}^{t(S_m-S_{m-1})}\Big|\mathbf{Z}_1^{m-1}\right\}\\
=&\mathrm{e}^{-t\xi}\prod_{m=1}^{N}\mathrm{E}\left\{\mathrm{e}^{t\psi_m}\right\}.
\end{aligned}$$

This completes the proof. ■

## 6.2 Proofs of Theorem 3.1 and Corollary 3.2

We need the following result presented in (Houdré, 2002).

**Lemma 6.2** *Let* $\mathbf{z}$ *be drawn from an ID distribution with the generating triplet* $(\mathbf{a},\mathbf{0},\nu)$ *such that* $\mathrm{E}\|\mathbf{z}\|^2<+\infty$. *If* $f,g:\mathbb{R}^K\rightarrow\mathbb{R}$ *are Lipschitz continuous functions, then*

$$\begin{aligned}\mathrm{E}f(\mathbf{z})g(\mathbf{z})-\mathrm{E}f(\mathbf{z})\mathrm{E}g(\mathbf{z})=\int_0^1\mathrm{E}_z\int_{\mathbb{R}^K}\big(f(\mathbf{z}+\mathbf{u})\\
-f(\mathbf{z})\big)\big(g(\mathbf{w}+\mathbf{u})-g(\mathbf{w})\big)\nu(d\mathbf{u})dz,\end{aligned}$$

*where the expectation* $\mathrm{E}_z$ *is signified in Lemma 1 of (Houdré, 2002).*

In the proofs of Theorem 3.1 and Corollary 3.2, we adopt some techniques used in Houdré's work (Houdré, 2002).

**Proof of Theorem 3.1.** First, we consider the validity of our proof. According to Theorem 25.3 in (Sato, 2004), we have

$$\begin{aligned}
\Omega=&\left\{t\geq 0:\mathrm{E}\mathrm{e}^{t\|\mathbf{z}\|}<+\infty\right\}\\
=&\left\{t\geq 0:\int_{\|\mathbf{u}\|>1}\mathrm{e}^{t\|\mathbf{u}\|}\nu(d\mathbf{u})<+\infty\right\}\\
=&\left\{t\geq 0:\int_{\|\mathbf{u}\|>1}\left(\mathrm{e}^{t\|\mathbf{u}\|}-t\|\mathbf{u}\|-1\right)\nu(d\mathbf{u})<+\infty\right\},
\end{aligned}$$

which implies that $\Omega$ is an interval and not reduced to $\{0\}$. Thus, the following discussion is valid.

Define $\mathbf{u}_1^N:=\{\mathbf{u}_n\}_{n=1}^N\subset\mathbb{R}^K$ and let $\mathbf{W}_1^N=\{\mathbf{w}_n\}_{n=1}^N$ be another sample set independently drawn from the ID distribution $\mathcal{Z}$. By combining (25), (26) and Lemma 6.2, for any $1\leq m\leq N$, we have

$$\begin{aligned}
&\mathrm{E}\left\{\psi_m(\mathbf{Z}_1^N)\mathrm{e}^{t\psi_m(\mathbf{Z}_1^N)}\right\}-\mathrm{E}\left\{\psi_m(\mathbf{Z}_1^N)\right\}\mathrm{E}\left\{\mathrm{e}^{t\psi_m(\mathbf{Z}_1^N)}\right\}\\
=&\int_0^1\mathrm{E}_z\Big\{\int\Big(\psi_m(\mathbf{Z}_1^N+\mathbf{u}_1^N)-\psi_m(\mathbf{Z}_1^N)\Big)\\
&\times\Big(\mathrm{e}^{t\psi_m(\mathbf{W}_1^N+\mathbf{u}_1^N)}-\mathrm{e}^{t\psi_m(\mathbf{W}_1^N)}\Big)\nu(d\mathbf{u}_1^N)\Big\}dz\\
=&\int_0^1\mathrm{E}_z\Big\{\mathrm{e}^{t\psi_m(\mathbf{W}_1^N)}\int\Big(\psi_m(\mathbf{Z}_1^N+\mathbf{u}_1^N)-\psi_m(\mathbf{Z}_1^N)\Big)\\
&\times\Big(\mathrm{e}^{t\left(\psi_m(\mathbf{W}_1^N+\mathbf{u}_1^N)-\psi_m(\mathbf{W}_1^N)\right)}-1\Big)\nu(d\mathbf{u}_1^N)\Big\}dz\\
\leq&\int_0^1\mathrm{E}_z\Big\{\mathrm{e}^{t\psi_m(\mathbf{W}_1^N)}\Big\}\int_{\mathbb{R}^K}\lambda\|\mathbf{u}_m\|\Big(\mathrm{e}^{t\lambda\|\mathbf{u}_m\|}-1\Big)\nu(d\mathbf{u}_m)dz\\
=&\mathrm{E}\Big\{\mathrm{e}^{t\psi_m(\mathbf{W}_1^N)}\Big\}\int_{\mathbb{R}^K}\lambda\|\mathbf{u}_m\|\Big(\mathrm{e}^{t\lambda\|\mathbf{u}_m\|}-1\Big)\nu(d\mathbf{u}_m).
\end{aligned}$$

Since the marginal distribution of $(\mathbf{Z}_1^N,\mathbf{W}_1^N)$ is $\mathbf{Z}_1^N$ and $\mathbf{Z}_1^N$ has the same distribution as that of $\mathbf{W}_1^N$, letting $L(t)=\mathrm{E}\mathrm{e}^{t\psi_m(\mathbf{W}_1^N)}$, we have

$$\begin{aligned}
\frac{L'(t)}{L(t)}=&\frac{\mathrm{E}\psi_m\mathrm{e}^{t\psi_m(\mathbf{Z}_1^N)}}{\mathrm{E}\mathrm{e}^{t\psi_m(\mathbf{Z}_1^N)}}\\
\leq&\int_{\mathbb{R}^K}\lambda\|\mathbf{u}_m\|\Big(\mathrm{e}^{t\lambda\|\mathbf{u}_m\|}-1\Big)\nu(d\mathbf{u}_m). \quad (28)
\end{aligned}$$

Therefore, we have

$$\int_0^t\frac{L'(s)}{L(s)}ds\leq\int_0^t\int_{\mathbb{R}^K}\lambda\|\mathbf{u}_m\|\Big(\mathrm{e}^{s\lambda\|\mathbf{u}_m\|}-1\Big)\nu(d\mathbf{u}_m)ds,$$

and then by (26),

$$\begin{aligned}
&\ln\mathrm{E}\mathrm{e}^{s\psi_m}\Big|_0^t=\ln\mathrm{E}\mathrm{e}^{t\psi_m}\\
\leq&\int_{\mathbb{R}^K}\Big(\mathrm{e}^{t\lambda\|\mathbf{u}_m\|}-t\lambda\|\mathbf{u}_m\|-1\Big)\nu(d\mathbf{u}_m). \quad (29)
\end{aligned}$$

By combining (29) and Lemma 6.1, we have

$$\Pr\left\{F\left(\mathbf{Z}_1^N\right)-\mathrm{E}F>\xi\right\}\leq\mathrm{e}^{\Phi(t)-t\xi}, \quad (30)$$

where

$$\Phi(t)=N\int_{\mathbb{R}^K}\Big(\mathrm{e}^{t\lambda\|\mathbf{u}\|}-t\lambda\|\mathbf{u}\|-1\Big)\nu(d\mathbf{u}). \quad (31)$$

Since $\mathrm{E}\mathrm{e}^{t\|\mathbf{z}\|}<+\infty$, for all $0<t<M$, $\Phi$ is infinitely differentiable on $(0,M)$ with

$$\Phi'(t)=\tau(t)=N\int_{\mathbb{R}^K}\lambda\|\mathbf{u}\|\Big(\mathrm{e}^{t\lambda\|\mathbf{u}\|}-1\Big)\nu(d\mathbf{u})>0, \quad (32)$$

and

$$\Phi''(t) = N \int_{\mathbb{R}^K} \lambda^2 \|\mathbf{u}\|^2 \mathrm{e}^{t\lambda\|\mathbf{u}\|} \nu(d\mathbf{u}) > 0. \qquad (33)$$

Then, we minimize the right-hand side of (30) with respect to $t$. According to (32) and (33), for any $0 < \xi < \tau(M^{-1})$, $\min_{0<t<M} \{\Phi(t) - t\xi\}$ is achieved when $\tau(t) - \xi = 0$. Since $\Phi(0) = \tau(0) = \tau^{-1}(0) = 0$, we have

$$\begin{aligned} \Phi\left(\tau^{-1}(\xi)\right) = & \int_0^{\tau^{-1}(\xi)} \tau(s) ds = \int_0^{\xi} s d\tau^{-1}(s) \\ = & \xi\tau^{-1}(\xi) - \int_0^{\xi} \tau^{-1}(s) ds. \end{aligned} \qquad (34)$$

Thus, for any $0 < \xi < \tau(M^{-1})$,

$$\min_{0<t<M} \{\Phi(t) - t\xi\} = - \int_0^{\xi} \tau^{-1}(s) ds.$$

Similarly, we also can prove that

$$\Pr\left\{\mathrm{E}F - F\left(\mathbf{Z}_1^N\right) > \xi\right\} \leq \exp\left(-\int_0^{\xi} \tau^{-1}(s) ds\right).$$

This completes the proof. ■

Next, we prove Corollary 3.2.

**Proof of Corollary 3.2.** Since $\nu$ has the bounded support $supp(\nu) \subseteq [-R, R]$, $\mathrm{E}\mathrm{e}^{t\|\mathbf{z}\|} < +\infty$ holds for any $t > 0$. Then, we have

$$\begin{aligned} \tau(t) = & N \int_{\|\mathbf{u}\| \leq R} \lambda \|\mathbf{u}\| \left(\mathrm{e}^{t\lambda\|\mathbf{u}\|} - 1\right) \nu(d\mathbf{u}) \\ = & N \int_{\|\mathbf{u}\| \leq R} \lambda^2 \|\mathbf{u}\|^2 \left(\sum_{k=1}^{\infty} \frac{t^k \lambda^{k-1} \|\mathbf{u}\|^{k-1}}{k!}\right) \nu(d\mathbf{u}) \\ \leq & N \int_{\|\mathbf{u}\| \leq R} \lambda^2 \|\mathbf{u}\|^2 \left(\sum_{k=1}^{\infty} \frac{t^k (\lambda R)^{k-1}}{k!}\right) \nu(d\mathbf{u}) \\ = & NV \left(\frac{\mathrm{e}^{t\lambda R} - 1}{\lambda R}\right). \end{aligned} \qquad (35)$$

As shown in (32) and (33), $\tau(t)$ is an increasing function and thus $\tau^{-1}(t)$ is also an increasing function. Moreover, according to Theorem 3.1 and (35), we have for any $\xi > 0$,

$$\begin{aligned} & \Pr\left\{F(\mathbf{Z}_1^N) - \mathrm{E}F > \xi\right\} \\ \leq & \exp\left\{-\int_0^{\xi} \frac{1}{\lambda R} \ln\left(1 + \frac{\lambda R s}{NV}\right) ds\right\} \\ = & \exp\left\{\frac{\xi}{\lambda R} - \left(\frac{\xi}{\lambda R} + \frac{NV}{\lambda^2 R^2}\right) \ln\left(1 + \frac{\xi\lambda R}{NV}\right)\right\}. \end{aligned} \qquad (36)$$

This completes the proof. ■

### 6.3 Proof of Theorem 4.2

Before the formal proof, we introduce the symmetrization inequality and details are given in (Bousquet *et al.*, 2004).

**Lemma 6.3** *(Symmetrization) Assume that $\mathcal{F}$ is a function class with the range $[A, B]$ and let $\mathbf{Z}_1^N, \mathbf{Z}'^N_1$ be two i.i.d. sample sets. Then, for any $\xi > 0$ such that $N\xi^2 \geq 32 \max\{A^2, B^2\}$, we have*

$$\begin{aligned} & \Pr\left\{\sup_{f \in \mathcal{F}} \left|\mathrm{E}f - \mathrm{E}_N f\right| > \xi\right\} \\ \leq & 2\Pr\left\{\sup_{f \in \mathcal{F}} \left|\mathrm{E}'_N f - \mathrm{E}_N f\right| > \frac{\xi}{2}\right\}. \end{aligned} \qquad (37)$$

Next, we prove Theorem 4.2.

**Proof of Theorem 4.2.** Consider $\{\epsilon_n\}_{n=1}^N$ as independent Rademacher random variables, *i.e.*, independent $\{-1, 1\}$-valued random variables with equal probability of taking either value. Given an $\{\epsilon_n\}_{n=1}^N$ and a $\mathbf{Z}_1^{2N}$, for any $f \in \mathcal{F}$, denote

$$\overrightarrow{\epsilon} := (\epsilon_1, \cdots, \epsilon_N, -\epsilon_1, \cdots, -\epsilon_N)^T, \qquad (38)$$

and

$$\overrightarrow{f}(\mathbf{Z}_1^{2N}) := \left(f(\mathbf{z}'_1), \cdots, f(\mathbf{z}'_N), f(\mathbf{z}_1), \cdots, f(\mathbf{z}_N)\right)^T.$$

According to Lemma 6.3, for any $\xi > 0$ such that $N\xi^2 \geq 32 \max\{A^2, B^2\}$, we have

$$\begin{aligned} & \Pr\left\{\sup_{f \in \mathcal{F}} \left|\mathrm{E}f - \mathrm{E}_N f\right| > \xi\right\} \\ \leq & 2\Pr\left\{\sup_{f \in \mathcal{F}} \left|\mathrm{E}'_N f - \mathrm{E}_N f\right| > \frac{\xi}{2}\right\} \\ = & 2\Pr\left\{\sup_{f \in \mathcal{F}} \left|\frac{1}{N} \sum_{n=1}^N \left(f(\mathbf{z}'_n) - f(\mathbf{z}_n)\right)\right| > \frac{\xi}{2}\right\} \\ = & 2\Pr\left\{\sup_{f \in \mathcal{F}} \left|\frac{1}{N} \sum_{n=1}^N \epsilon_n \left(f(\mathbf{z}'_n) - f(\mathbf{z}_n)\right)\right| > \frac{\xi}{2}\right\} \\ = & 2\Pr\left\{\sup_{f \in \mathcal{F}} \left|\frac{1}{2N} \langle \overrightarrow{\epsilon}, \overrightarrow{f}(\mathbf{Z}_1^{2N}) \rangle\right| > \frac{\xi}{4}\right\}. \end{aligned} \qquad (39)$$

Set $\Lambda$ to be an $\xi/8$-radius cover of $\mathcal{F}$ with respect to the $L_1(\mathbf{Z}_1^{2N})$ norm. Since $\mathcal{F}$ is composed of Lipschitz continuous functions, we assume that the same holds for any $h \in \Lambda$. If $f$ is the function that achieves $\sup_{f \in \mathcal{F}} \frac{1}{2N} \left|\langle \overrightarrow{\epsilon}, \overrightarrow{f}(\mathbf{Z}_1^{2N}) \rangle\right| > \frac{\xi}{4}$, there must be an $h \in \Lambda$ that satisfies that

$$\frac{1}{2N} \sum_{n=1}^N \left(|f(\mathbf{z}'_n) - h(\mathbf{z}'_n)| + |f(\mathbf{z}_n) - h(\mathbf{z}_n)|\right) < \frac{\xi}{8},$$

and

$$\sup_{h \in \Lambda} \frac{1}{2N} \left|\langle \overrightarrow{\epsilon}, \overrightarrow{h}(\mathbf{Z}_1^{2N}) \rangle\right| > \frac{\xi}{8}.$$

Therefore, by (39), we have

$$
\begin{aligned}
&\Pr\left\{\sup_{f\in\mathcal{F}}\left|\mathrm{E}f-\mathrm{E}_N f\right|>\xi\right\}\\
\leq&2\Pr\left\{\sup_{f\in\mathcal{F}}\left|\frac{1}{2N}\langle\overrightarrow{\epsilon},\overrightarrow{f}(\mathbf{Z}_1^{2N})\rangle\right|>\frac{\xi}{4}\right\}\\
\leq&2\Pr\left\{\sup_{h\in\Lambda}\left|\frac{1}{2N}\langle\overrightarrow{\epsilon},\overrightarrow{h}(\mathbf{Z}_1^{2N})\rangle\right|>\frac{\xi}{8}\right\}\\
\leq&2\mathrm{E}\left\{\mathcal{N}\left(\mathcal{F},\frac{\xi}{8},L_1(\mathbf{Z}_1^{2N})\right)\right\}\\
&\times\max_{h\in\Lambda}\Pr\left\{\left|\mathrm{E}'_N h-\mathrm{E}_N h\right|>\frac{\xi}{4}\right\}\\
\leq&2\mathrm{E}\left\{\mathcal{N}\left(\mathcal{F},\frac{\xi}{8},L_1(\mathbf{Z}_1^{2N})\right)\right\}\\
&\times\max_{h\in\Lambda}\Pr\left\{|\mathrm{E}h-\mathrm{E}'_N h|+|\mathrm{E}h-\mathrm{E}_N h|>\frac{\xi}{4}\right\}\\
=&2\mathrm{E}\left\{\mathcal{N}\left(\mathcal{F},\frac{\xi}{8},L_1(\mathbf{Z}_1^{2N})\right)\right\}\\
&\times\max_{h\in\Lambda}\Pr\left\{\left|\mathrm{E}h-\mathrm{E}_N h\right|>\frac{\xi}{8}\right\}.
\end{aligned}\tag{40}
$$

By combining Theorem 3.1 and (40), we can obtain Theorem 4.2. This completes the proof. ■

## 7 Conclusion

In this paper, we study the risk bounds for samples independently drawn from an infinitely divisible (ID) distribution with the generating triplet $(\mathbf{a},\mathbf{0},\nu)$. By using a martingale method, we provide two kinds of deviation inequalities for a sequence of ID random variables with $(\mathbf{a},\mathbf{0},\nu)$. We then utilize the resulted deviation inequalities to obtain the risk bounds based on the covering number. We present a sufficient condition for the asymptotic convergence of the risk bound (12) and it is in accordance with the result for the generic i.i.d. empirical process (*cf.* Theorem 2.3 in (Mendelson, 2003)). We further show that if samples are independently drawn from an ID distribution with $(\mathbf{a},\mathbf{0},\nu)$, the convergence rate of $\sup_{f\in\mathcal{F}}\left|\mathrm{E}f-\mathrm{E}_N f\right|$ reaches $O\big(\big(\frac{\ln \mathrm{E}\{\mathcal{N}(\mathcal{F},\xi/8,L_1(\mathbf{Z}_1^{2N}))\}}{N}\big)^{\frac{1}{1.3}}\big)$ and it is faster than $O\big(\big(\frac{\ln \mathrm{E}\{\mathcal{N}(\mathcal{F},\xi,L_1(\mathbf{Z}_1^{N}))\}}{N}\big)^{\frac{1}{2}}\big)$ for the generic i.i.d. empirical process (Mendelson, 2003).

### Acknowledgements

This work was supported in part by Australian Research Council Future Fellowship (Grant No. FT100100971), Australian Research Council (ARC) Discovery Project DP1093762, Discovery Project DP0988016 and the Nanyang Technological University Nanyang SUG Grant (under project number M58020010).